\documentclass{article}
\usepackage{amsmath}
\usepackage[a4paper, total={6in, 8in}]{geometry}
\usepackage{authblk}

\usepackage[utf8]{inputenc}
\usepackage{graphicx}
\usepackage{xcolor}
\usepackage{listings}

\begin{document}
\title{A human factors approach to validating driver models for interaction-aware automated vehicles}
\author[1]{Olger Siebinga}
\author[1]{Arkady Zgonnikov}
\author[1]{David Abbink}
\affil[1]{Delft University of Technology, the Netherlands}
\renewcommand\Affilfont{\itshape\small}
\date{September 2021}

\maketitle
\begin{abstract}
A major challenge for autonomous vehicles is interacting with other traffic participants safely and smoothly. A promising approach to handle such traffic interactions is equipping autonomous vehicles with interaction-aware controllers (IACs). These controllers predict how surrounding human drivers will respond to the autonomous vehicle's actions, based on a driver model. However, the predictive validity of driver models used in IACs is rarely validated, which can limit the interactive capabilities of IACs outside the simple simulated environments in which they are demonstrated. In this paper, we argue that besides evaluating the interactive capabilities of IACs, their underlying driver models should be validated on natural human driving behavior. We propose a workflow for this validation that includes scenario-based data extraction and a two-stage (tactical/operational) evaluation procedure based on human factors literature. We demonstrate this workflow in a case study on an inverse-reinforcement-learning-based driver model replicated from an existing IAC. This model only showed the correct tactical behavior in $40\%$ of the predictions. The model's operational behavior was inconsistent with observed human behavior. The case study illustrates that a principled evaluation workflow is useful and needed. We believe that our workflow will support the development of appropriate driver models for future automated vehicles.
\end{abstract}

\section{Introduction}
\begin{figure}[h!]
    \centering
    \includegraphics[width=\textwidth]{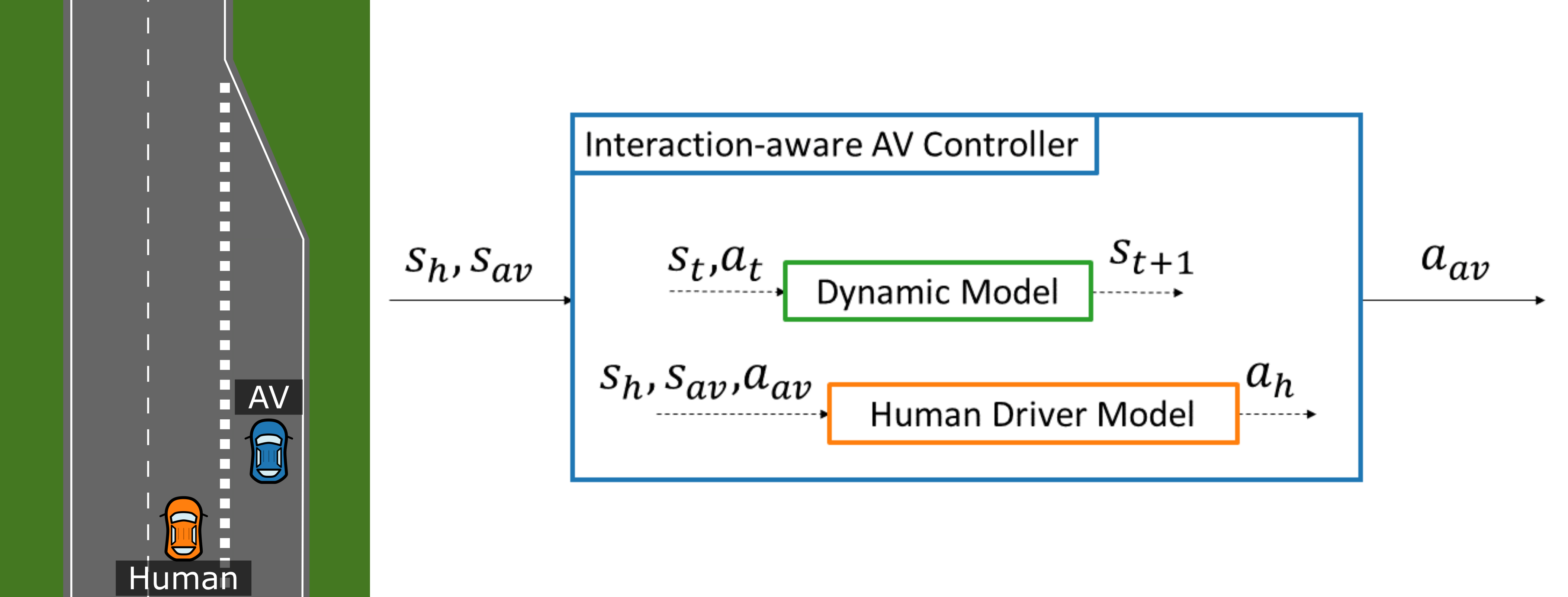}
    \caption{A high-level diagram of a typical interaction-aware controller (IAC) for autonomous vehicles (AVs). Such a controller operates in situations where the states and actions of a human-driven vehicle (superscript $h$) and an AV (superscript $av$) influence each other, e.g. the merging situation depicted in the left panel. Future states and actions are denoted with subscript $t+1$, all other states and actions are at time $t$. An IAC determines the optimal action $a$ for the AV based on the current state $s$ of both the AV and the human. To find this optimal action, IACs make use of at least two prediction models: a dynamic model to predict future states ($s^*_{t+1}$) based on current states ($s_t^*$) and actions ($a_t^*$)(the superscript $*$ denotes it can either be used for the AV or the human)}, and a human driver model to predict the actions surrounding human drivers will take in response to the AV's action. Both the dynamic and human behavior predictions are evaluated to find the optimal action for the AV, this is usually done with a reward function that incorporates aspects like safety and comfort. The validation of the human driver model is the focus of this work.
    \label{fig:iac_block_diagram}
\end{figure}

One of the great technological and societal promises of the 21st century is the autonomous vehicle (AV)~\cite{Harper2016, Clements2017, Pettigrew2017}. This technology has been under development in laboratories and under controlled conditions for decades and is now transitioning to the real world. However, a major challenge for real-world implementation of AV technologies is enabling AVs to handle complex interactions with human road users. AV controllers have recently been proposed that aim to address this challenge through \textit{interaction-aware} controllers (IACs)~\cite{Sadigh2018, Schwarting2019, Evestedt2016, Ward2017, Liu2015, Meng2016, Lenz2016, Tian2019, Zhang2018, Yu2018, Hubmann2018, Coskun2019, Garzon2019, Isele2019, Hang2021}. IACs incorporate a model of human driver behavior in the controller, to predict how another driver is likely to respond to the AV’s behavior. Based on this prediction and its own reward function (e.g., incorporating safety, comfort, etc.), the IAC finds the optimal action for the AV (Figure~\ref{fig:iac_block_diagram}). However, up to now the interactive capabilities of these controllers have only been demonstrated in simplified simulated environments (e.g. top-down view computer simulations). Whether the state-of-the-art IACs are capable of predicting naturalistic driver behavior and interacting with humans in real traffic remains an open question.

Although demonstrating a proposed controller in a simulated traffic environment is a necessary first step to show its potential, it does not provide sufficient evidence on how well the controller will generalize to real-world environments. In this work, we take the position that before implementing an IAC in vehicles to validate its behavior in the real world, its underlying driver model should be validated on natural human driving behavior. If the model fails to predict real-world behavior accurately, the controller will act on false predictions which can lead to annoying or even unsafe situations. Such driver model validation can therefore provide an early indication of the IAC validity without much of the cost associated with implementing and testing it in real traffic interactions. However, driver model validation is currently not a part of the mainstream approach to IAC validation (see e.g.~\cite{Sadigh2018, Schwarting2019, Ward2017}), and a principled framework for such validation is missing from the literature.

The contribution of our work lies in proposing and demonstrating a human-factors-based evaluation workflow, in order to help IAC designers in the process of selecting appropriate driver models. The proposed workflow validates driver models using empirical data obtained from \textit{naturalistic} (real-world) traffic interactions, acknowledging two levels of driving behavior~\cite{Michon1985}: \textit{tactical} choices and \textit{operational} safety margins. Tactical behavior refers to \textit{which} maneuvers are executed (e.g., a lane change or car following) and operational behavior describes \textit{how} they are executed (e.g., in terms of safety margins). To demonstrate the potential of this workflow, we perform a case study that shows that an inverse-reinforcement-learning-based model, replicated from a model used in a previously developed IAC~\cite{Sadigh2018}, does not generalize to real-world data. Even though we do not quantify the implications of these results for any specific IAC, they still underline the importance of using validated driver models in AV controllers.

\section{Validating driver models for interaction-aware controllers}

\subsection{Why validate?}
Part of the reason why model validation is necessary is that the simulated environments in which IACs are evaluated are not sufficient to assume safe generalization to the real world. A particular aspect of the evaluation is the human response to the AV's actions. Two approaches to generate this response are used. Some studies~\cite{Schwarting2019, Evestedt2016, Ward2017, Liu2015, Meng2016, Lenz2016, Tian2019, Zhang2018, Garzon2019, Isele2019, Hang2021} simulate human driver responses using driver models. However, many of the driver models used for this purpose are also not validated on natural human driver behavior, which could indicate a discrepancy between the simulation and natural behavior. Other studies~\cite{Sadigh2018, Hubmann2018, Coskun2019, Yu2018} use real-time responses of a human test subject in an abstract top-down view computer simulation, much like a video game. The gap between such abstract test environments and real-world driving is large, e.g. due to the absence of risk perception~\cite{Ranney2011}, motion cues, and visual looming~\cite{Lee1976}. So, again we can expect the participants' responses to differ from driver responses in real-world traffic. This means that both approaches can only provide very limited evidence for generalization of the demonstrated interactive capabilities of the IAC to the real world. 

To show that the IAC's behavior does generalize to the real world, one could propose to implement the IAC in a real vehicle and demonstrate its workings in a natural environment. However, deploying a proof-of-concept IAC in the real world might result in unsafe situations even under highly controlled conditions. This raises ethical concerns about such real-world testing. Another possibility would be to use real-time human responses and minimize the mismatch between the simulation environment and the real world, e.g. by using a high-fidelity driving simulator. However, such experiments are expensive and time-consuming, and human behavior even in realistic driving simulators can still differ from behavior in real traffic~\cite{Ranney2011, Greenberg2011}. For this reason, we advocate a complementary approach: validating the driver model on naturalistic traffic data before implementing it in an IAC. The combination of the model validation on real-world data and demonstrating the IAC's interactive capabilities in a (simplified) simulated environment provides a firm ground for the further implementation and testing of the IAC in real vehicles.

To the best of our knowledge, validation on naturalistic driving data for use in IACs has not been performed for two of the most commonly used driver models proposed for IACs. These models are the intelligent driver model IDM~\cite{Treiber2000} (used in~\cite{Evestedt2016, Ward2017, Hubmann2018} to predict driver behavior and in~\cite{Isele2019, Coskun2019, Zhang2018} to simulate other drivers' responses) and the expected-utility-maximizing model (used e.g. in~\cite{Schwarting2019, Sadigh2018} to predict other drivers' behavior) that uses a reward function learned from human demonstrations with inverse reinforcement learning (IRL). Although the reward function in this model is learned from naturalistic driving data, none of the studies which proposed IACs based on an IRL-based model have validated the resulting model with respect to its ability to capture human behavior.

\subsection{How to validate?}
\label{sec:how_to_validate}
\begin{figure}
    \centering
    \includegraphics[width=\textwidth]{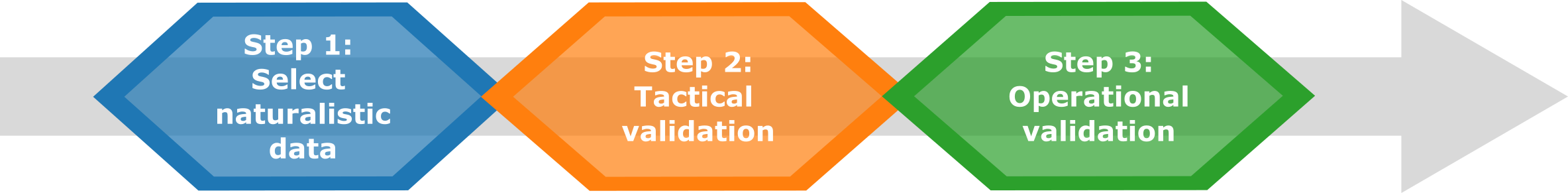}
    \caption{The proposed driver model validation workflow for interaction-aware autonomous vehicle controllers. The workflow consists of three steps. In the first step, a suitable dataset is selected to perform the validation of the driver model on. From this selected dataset, specific situations are automatically extracted. The actual validation of the model takes place in the last two steps. A distinction is made based on the level of behavior. First, the tactical behavior is validated in step 2. This step reveals to what extent the driver model shows tactical behavior that is consistent with human behavior in the dataset. Behavior inconsistent with human data, e.g. collisions, is not regarded in the final step. The third step evaluates the operational behavior of the model based on human factors literature. This is done for every tactical behavior separately. The final conclusion of the validation should be based on the combined results of steps 2 and 3.}
    \label{fig:proposed_workflow}
\end{figure}

We propose a three-step evaluation workflow (Figure~\ref{fig:proposed_workflow}) that incorporates important aspects of driver model validation: evaluation against naturalistic data on both the tactical and operational levels. 

\subsubsection*{Step 1: Select naturalistic data}
When validating a driver model for an IAC, we propose that the model is compared against human behavior data recorded in a natural environment, i.e. a naturalistic driving dataset. There are increasingly many naturalistic datasets available, but which dataset should one choose? And once the dataset is chosen, should all the data in the dataset be used uniformly for model validation?

When selecting a naturalistic dataset, one should be aware whether the data recording was done with obtrusive or unobtrusive methods. Obtrusive methods are methods where the driver is aware their behavior is being recorded (e.g., the SHRP2 dataset~\cite{Antin2019}). As a result, the driver might have changed their behavior e.g., to conform to the expectations of the researchers. Other datasets are gathered \textit{without} the drivers knowing that their behavior is being recorded, typically with drones and cameras (several open-access datasets are available e.g., \cite{Krajewski2018, NGSIM2016, Barmpounakis2020}). Because of the possibility of adapted behavior in obtrusive naturalistic datasets, unobtrusive datasets are preferable for model validation.  

When a suitable dataset is chosen, specific parts of the data need to be selected to perform the validation on. Data recorded in the real world often contains many different scenarios, e.g. different locations, vehicle types, and maneuvers. Using all this data to validate a driver model would be intractable because humans behave differently in different scenarios. Instead, comparable scenarios can be selected from the dataset to be evaluated together. These scenarios should fit the intended environment of the IAC. At the same time, one should avoid hand-picking scenarios, or selecting them on low-level characteristics (e.g. only include vehicles that reach a certain velocity) because this will reduce the variability in the data and thus negate the purpose of the validation, to show that the model generalizes to real-world behaviors. Instead, scenarios should be selected on higher-level similarities, e.g. include \textit{all} lane changes or \textit{all} unprotected left turns. Open-source software is available that includes examples of how to extract such scenarios automatically e.g.,~\cite{Siebinga2021}. 

\subsubsection*{Behavior validation}
After selecting relevant scenarios, the model can be trained and validated. Validation of models of human behavior is often difficult because there are many aspects that determine if the model's behavior resembles human behavior. In most cases, the difference cannot be captured by a single metric. For example: when validating a driver model in a lane-changing scenario, it could be tempting to use a distance-based error-metric to describe the goodness-of-fit. However, an event like a collision with a vehicle in an adjacent lane can, in some cases, be described by a small lateral distance error with respect to a human-driven trajectory. If only this distance error would be examined when validating the model, it would seem to perform well, but in reality, the model predicts that a human would collide with another vehicle. The collision is missed in the single-metric validation procedure, and the (wrong) conclusion would be that the model describes human behavior with only a small error margin.

This example illustrates that a distinction should be made between \textit{what} behavior is executed (e.g., car following, crashing, or lane changing) and \textit{how} it is executed (i.e. specific trajectories and safety margins with respect to lane boundaries and traffic participants). This bears resemblance to the common distinction in driving behavior~\cite{Michon1985} of tactical and operational behavior (note that strategic behavior, e.g. route selection, is not covered by the models in IACs). In this distinction, the maneuvers executed by the driver, like a lane change, are tactical behavior. The manner in which they are executed, e.g., expressed in accelerations or dynamics of the gaps with respect to other vehicles, is called operational behavior. Making this distinction in driver model validation is especially relevant for driver models used in IACs because these models are mostly designed to incorporate multiple tactical behaviors. This is in contrast to traditional driver models that were more often designed to only represent one specific tactical behavior.

For many tactical behaviors, the corresponding operational behavior has been studied in human-factors experiments (e.g., for car following~\cite{Saifuzzaman2014,Ossen2011,Hoogendoorn2006,Jiang2015,Mulder2005,KONDOH2008}). These studies provide the important metrics of human operational behavior, given a specific tactical behavior. Making the same distinction during the validation allows one to leverage the existing human factors literature, enabling researchers without in-depth human-factors expertise to validate their models. 

Determining what tactical behavior is executed by the model and if it matches human behavior is something that can be done without any expert knowledge. For instance, it is straightforward to specify if a lane change is made, and to compare if the model performs a lane change in the same situation where a human does. Once the tactical behavior is determined, the metrics specifying the operational behavior can be defined based on the relevant human-factors literature. This will require obtaining some knowledge on the subject, but with a properly specified tactical behavior, a brief, non-exhaustive driver-behavior literature survey would be enough for a researcher to make a motivated choice of the metrics characterizing the corresponding operational behavior.

Because making a distinction between tactical and operational behavior is relevant for IACs and makes the validation process easier, we propose a sequential two-stage validation process. The first stage (step 2 in the workflow of Figure~\ref{fig:proposed_workflow}) is to validate the model's behavior on a tactical level, providing a quick and straightforward distinction between behavior that clearly resembles or does not resemble the observed human driving behavior in the same circumstances. The second stage (step 3 in Figure~\ref{fig:proposed_workflow}) examines the tactical behaviors separately on the operational level.

\subsubsection*{Step 2: Tactical validation}
The purpose of the tactical validation step is two-fold. First, it serves to determine which of the model's responses are consistent with human behavior and which are not. A valid driver model does not predict tactical responses inconsistent with human behavior, therefore we will refer to such responses as undesirable tactical behavior. Desirable behaviors on the other hand, are all tactical responses that can be observed in naturalistic human driving data. Second, this step will categorize the model's responses so its desirable behaviors can be validated in the operational validation step according to the right criteria. Undesirable behavior can be disregarded during the operational validation step because it does not matter how the model performs a behavior that is undesirable in the first place.

To achieve this, a mutually exclusive set of \textit{possible} tactical behaviors exhibited by the model should be defined. The distinction between these tactical behaviors should be based on simple rules (or inclusion and exclusion criteria) such that all exhibited model behavior falls in one and only one tactical category. Which and how many of these categories to include depends on the outcome of the literature survey discussed earlier. All behaviors in one category should be validated on the same operational characteristics, which should be taken into account when determining the categories.

\subsubsection*{Step 3: Operational validation}
For the operational validation step, human-factors literature provides signals and metrics that best describe human behavior for specific tactical behavior. This operational validation step can compare individual trajectories or averaged metrics between human and model behavior as long as the metrics and signals are chosen appropriately and the tactical behaviors are regarded separately. Examples of such metrics are metrics that relate to the dynamics of the behavior, e.g. the gap between vehicles, or to the properties of the maneuver, e.g. the duration of a lane change. Human-factors literature can also provide methods on \textit{how} to compare the signals and metrics. For example, in~\cite{Saifuzzaman2014} figures are presented that relate phase diagrams in car following to responsive actions of human drivers, such plotting methods can also be used for model validation.

\subsubsection*{The validation conclusion}
The final conclusion of the validation procedure should be based on both the tactical and operational behavior displayed by the model. The model should display desirable tactical behavior in a way that resembles how humans perform the same behavior on an operational level. But because the eventual goal is to incorporate the driver model in an IAC, the controller's ability to safely operate while using the model's predictions can be seen as the most important factor in the final conclusion. 

When a driver model shows behavior that deviates from human behavior to a large extent, but the controller that implements the model can still safely operate with these errors, it can still be concluded that the model is "good enough" for use in the IAC. To draw such a conclusion, the maximal acceptable difference between the model's output and human behavior has to be defined. This should be done for every IAC separately due to differences in IACs, scenarios, and regarded tactical behaviors. The maximal acceptable difference can for example be based on an evaluation that shows that the controller can still reliably execute safe and acceptable interactive behavior when confronted with predictions that have this maximal deviation from future human behavior. 

However, even if an IAC is robust to inaccurate predictions of the driver model, we argue that it is still important to validate the model and report the magnitude of the deviation from human behavior. This improves the re-usability of the proposed model for other IACs and provides a basis for a re-evaluation of the model when extending or improving the IAC.

\section{Case study: Methods}
To demonstrate the proposed workflow we use it to validate an inverse reinforcement learning (IRL) based model replicated from a study that proposed one of the first IACs for autonomous vehicles~\cite{Sadigh2018}. The choice to validate an IRL-based model was made because this increasingly popular type of model describes dynamic human behavior in multiple scenarios and has not been validated previously. The two IACs with IRL-based driver models discussed earlier~\cite{Sadigh2018, Schwarting2019} use similar implementations of such a model. However, only the work by Sadigh et al.~\cite{Sadigh2018} provides enough detail, in the form of mathematical description and open-source code, to replicate the used IRL-based model. For that reason, the model used by Sadigh et al. is used as a reference for this case study. 

\subsection{Model implementation}
\label{sec:implementation}
IRL-based driver models assume that human behavior is "driven" by an underlying reward function. A parameterized reward function is assumed and inverse reinforcement learning is used to infer the parameters directly from human demonstrations (see~\cite{Ng2000, Abbeel2004, Ziebart2008}). This reward function with the learned parameters can be used in an agent to generate individual predictions of human behavior. Driver models based on IRL use a utility-maximizing rational agent for this purpose. Throughout this paper, we refer to this method of generating predictions combined with a specific assumed reward function as the model. We refer to instances of the model with a specific set of parameters as an agent. In IRL-based driver models, the used reward function consists of a linear combination of features, each with its own weight:

\begin{equation}
\label{eq:reward_irl}
R^h(s, a) = \sum \theta_i^h \phi_i(s, a).
\end{equation}

In this formula, $R^h$ denotes the reward of a specific human, $s$ is the state (at time $t$) and $a$ is the action sequence the human will take. This action sequence is subject to a finite planning horizon. $\phi_i$ denotes the $i_{th}$ feature and $\theta_i$ represents the corresponding weight, which is learned by IRL from demonstrations produced by a human driver $h$. Note that the features $\phi_i$ in equation~\eqref{eq:reward_irl} are designed beforehand and do not vary over humans, demonstrations, or situations. The weights $\theta_i$ are learned from the demonstrations and vary over humans. These weights are learned by maximizing the log-likelihood of an observed demonstration with respect to the weights, given the assumed features. 

\subsection{Assumed reward function}
The reward function $R^h$ used for the IRL-based model in this work was replicated from~\cite{Sadigh2018} and consists of four features for: maintaining a preferred velocity, lane-keeping, staying on the road, and collision avoidance. The collision avoidance feature is modeled by a two-dimensional Gaussian function, based on distances between the centers of vehicles. Because the human demonstrations we use for the case study were recorded on highways, the heading angles of the vehicles take very low values and are therefore neglected for collision avoidance. They are assumed to be equal to the road heading (this is a deviation from the model used in~\cite{Sadigh2018}). The lane-keeping and road boundary features are both Gaussian functions of the lateral road axis, they are constant over the longitudinal axis of the road. The velocity feature is the squared error with respect to the desired velocity. Since the exact desired velocity is not known for the human drivers that provide the demonstrations, and the legal speed limits that could be used for this purpose are not always provided with the data, the maximum recorded velocity of a vehicle is taken as the driver's desired velocity. The full reward function is given in equation~\eqref{eq:reward_ours}.

\begin{equation}
\label{eq:reward_ours}
R^h(x, y, v_x) = \theta_\text{vel}^h \phi_\text{vel}(v_x) + \theta_\text{lane}^h \phi_\text{lane}(y) + \theta_\text{bounds}^h \phi_\text{bounds}(y) + \theta_\text{collision}^h \phi_\text{collision}(x, y),
\end{equation}

where

\[\phi_\text{vel}(v_x) = (v_x - v_d) ^ 2 \]
\[\phi_\text{lane}(y) =  e^{-c  (y_{lc} - y) ^2}\]
\[\phi_\text{bounds}(y) =  e^{-c  (y_{rb} - y) ^2}\]
\[ \phi_\text{collision}(x, y) = \frac{1}{\sigma_x  \sqrt{2\pi}}  e^{-(1/2)  ((x - x_o)^2/\sigma_x^2)} \frac{1}{\sigma_y  \sqrt{2\pi}}  e^{-(1/2)  ((y - y_o)^2/\sigma_y^2)}  \]

\begin{figure}[ht]
    \centering
    \includegraphics[width=\textwidth]{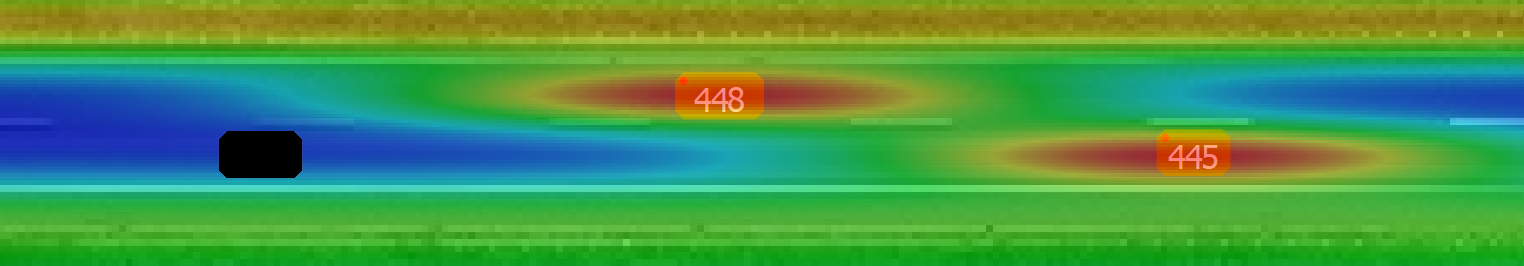}
    \caption{A heat map of the reward function (equation~\eqref{eq:reward_ours}) used for the IRL-based driver model where the black block indicates the ego vehicle. Warmer colors indicate low reward, cooler colors indicate high reward. The feature for velocity is not shown here because it does not depend on the position. The dimensions and positions of the features displayed here are assumed to be constant over different humans. The weights, represented here by the colors, differ between humans and are learned from demonstrations.}
    \label{fig:heatmap_example}
\end{figure}

In these formulae, $x$ and $y$ denote the longitudinal and lateral position as defined in Figure~\ref{fig:high_d_example}, $lc$ and $rb$ denote the lane center and road boundaries respectively, where the road boundaries are defined at half a lane width outside the outermost marking. $v$ represents velocity and subscript $o$ denotes the other vehicle. The constants $c$, $\sigma_x$ and $\sigma_y$ are used to shape the features. A visual representation of the reward function, excluding the velocity feature, can be found in Figure~\ref{fig:heatmap_example}.
 
The constants that shape these features were determined with a grid search on the first 15 demonstrations of the used dataset. Initial guesses of the parameters were based on a visual comparison of the heat map to the road image. Variations around these initial guesses were estimated based on dimensions of the lanes. (For example $min(\sigma_y) = 1.4~m$, thus $95.4\%$ of the lateral influence on collision prevention lies within $2.8~m$ distance between vehicle centers. With a lane width of $4~m$ and a $2~m$ wide vehicle, this means the lane marking has to be crossed before the collision prevention starts contributing to the reward. Thus, the lower bounds of our parameter grid are close to the smallest plausible parameter values.) We used the following sets in the grid search: $c=\{\textbf{0.14}, 0.18, 0.22\}, \sigma_x=\{5.0, 10.0,\textbf{15.0},20.0\}, \sigma_y=\{\textbf{1.4}, 1.8, 2.2\}$, where the \textbf{bold} value the selected value. Each parameter combination in the grid was evaluated based on the resulting number of desired tactical behaviors by the agent (see Section~\ref{sec:how_to_validate}, Step 2 for the definition of desired behavior). The parameter sets $c, \sigma_x, \sigma_y = 0.14, 15.0, 1.4$ and $c, \sigma_x, \sigma_y = 0.14, 20.0, 1.4$ had the maximum number of desired tactical behaviors in this grid search, we chose to select the combination containing our initial guess.

\subsection{Using the proposed workflow}
Here we will discuss the use of the proposed workflow (Figure~\ref{fig:proposed_workflow}) to validate the IRL-based model with the reward function as shown in equation~\eqref{eq:reward_ours} step by step.

\subsubsection{Step 1: select data}

The first step of the proposed workflow is to select a naturalistic dataset. Among multiple naturalistic driving datasets that are openly available, in this case study we considered three datasets: the NGSIM dataset~\cite{NGSIM2016}, the pNEUMA dataset~\cite{Barmpounakis2020}, and the HighD dataset~\cite{Krajewski2018}. Of these three the NGSIM data has larger uncertainties in the trajectories because it was recorded with fixed-base cameras instead of drones. The pNEUMA dataset was recorded in an urban environment, this does not match the environment of the regarded IAC~\cite{Sadigh2018} which focuses on multi-lane scenarios (e.g. a highway) with human behavior that mostly consists of actions to prevent collisions, like lane changing. The HighD dataset contains high-precision data recorded in a multi-lane environment. It also contains dynamic behavior such as lane changes to prevent collisions. For these reasons, we will use the HighD dataset. This dataset consists of 60 separate recordings, recorded in 6 different locations in Germany. All recordings were made on highways using drones equipped with cameras; from these recordings, trajectory data was automatically extracted~\cite{Krajewski2018}. Every recording is of a fixed stretch of highway, the average length of these recorded stretches is $416~m$, the average duration of a single-vehicle track is $14.34~s$.

\begin{figure}[ht]
    \centering
    \includegraphics[width=\textwidth]{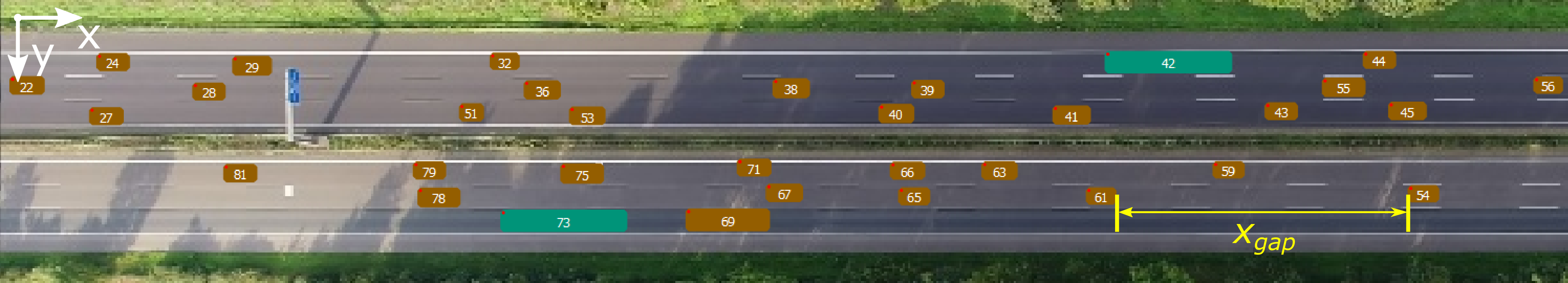}
    \caption{An example frame of the HighD dataset~\cite{Krajewski2018} as visualized using TraViA~\cite{Siebinga2021}. The frame includes a stretch of a highway in Germany, where vehicles drive on the right side of the road and where, in some of the cases, there are no legal speed limits. The orange shapes represent regular cars, the green shapes are trucks. All vehicles have a vehicle-ID shown in white. The white arrows display the coordinate frame and the yellow marking shows a visualization of the gap between two vehicles as used in the metrics for step 3 of the validation workflow.}
    \label{fig:high_d_example}
\end{figure}

To visualize the data and the resulting agent behavior we used TraViA~\cite{Siebinga2021}, an open-source visualization and annotation tool for trajectory datasets. TraViA can visualize all mentioned datasets and we extended it to train and visualize the IRL-based model. The extension code is available online~\cite{Siebinga2021b}. An example frame of the HighD dataset visualization can be found in Figure~\ref{fig:high_d_example}.

From this dataset, we automatically select suitable scenarios for training and validating the model. These scenarios should fit the intended use of the IAC~\cite{Sadigh2018}: in our case study, we assume the goal of the IAC is to interact with human drivers who perform lane changes. This means we could consider two distinct behaviors in the HighD dataset: lane changing and merging. A merging lane is only present in 3 of the 60 HighD recordings. For this reason, we will use human lane-changing maneuvers for validation. For consistency, the three recordings with a merging lane were not considered. 

As mentioned before, the features in the reward function consider collision avoidance, lane-keeping, staying on the road, and maintaining a preferred velocity. This means that not all lane changes can be explained with this model. Lane changes to the right are not covered because they are "driven" by a need to adhere to (socially acceptable) traffic rules that are not incorporated in the reward function (In Germany, it is obligatory to drive in the rightmost lane if it is free. So a lane change to the right is most often performed simply because that lane is free, not to avoid a collision. It can therefore not be explained by the used reward function). Therefore, only single lane changes to a left lane are considered for training and validation. The highD dataset includes the number of lane changes for every trajectory (based on lane crossings) and the current lane number at every frame. We automatically extracted all used trajectories based on these metrics.

\subsubsection{Step 2: tactical validation}
The next step is to define a set of tactical behavior categories. There are only a limited number of possible tactical behaviors on a highway without an exit lane, we will consider four possibilities: car following, lane changing, colliding, and crossing the road boundaries. Lane-keeping is not regarded as a separate behavior since all vehicles on a highway essentially follow another vehicle. In this set car following and lane changing are regarded as desirable behaviors, and colliding and going off-road are considered undesirable.

Besides defining the behavior categories, we established a procedure to place the trajectories produced by the agent in one of these categories based on a hierarchy in tactical behaviors. First, if an agent collided with another vehicle, this is  labeled as “collision”. If the agent did not collide, a check is done to see if the center of the vehicle stayed within the outer road boundaries; if not, the tactical behavior is labeled “off-road”. Agents that did not fall in one of the two categories above are checked for lane changes; if there is one, the tactical behavior is labeled “lane change”. And finally, agents that showed none of these three behaviors are placed in the “car following” category. All of these checks are implemented in the software and are performed automatically for all agents by checking for overlap with other vehicles and evaluating the vehicle's center position for every time step.

The used hierarchy is based on the idea that a predicted collision has the highest impact for IACs. If a model predicts a collision, an IAC will act to avoid this, independent of the fact that the model predicts a lane change first. Vehicles leaving the road will also have a big impact on IAC behavior because it reduces the number of vehicles to consider and thus changes the scene. However, the IAC will not take drastic actions to avoid this, therefore it comes second in the hierarchy. Only if none of these undesirable behaviors are executed by the model, lane changes are relevant. Finally, all other behaviors within a single lane is grouped as car following. A more fine-grained distinction could have been made here by including behaviors such as nudging or aborted lane changes. But before considering those more sophisticated behaviors, we chose to evaluate if and how the model displays car following in general.

To evaluate if the model's tactical performance is adequate for use in an IAC, a maximum acceptable deviation from human behavior needs to be specified. Because no IAC implementation is used in this case study, we cannot specify such a threshold here. 
 
\subsubsection{Step 3: operational validation}

The last step is to determine how to evaluate the model's operational behavior for the cases where the tactical behavior falls in one of the desirable categories. We have defined two desirable categories: lane changes and car following. Earlier studies investigated human car-following behavior and risk perception using inverse time-to-collision vs time gap plots~\cite{Mulder2009, KONDOH2008}, these metrics were also used to evaluate human lane changes before~\cite{Dang2013}.

Time-to-collision is defined as the time it will take until a vehicle collides with the preceding vehicle given that they both continue at their current velocity, 
\begin{equation}
\label{eq:ttc}
\text{TTC} = \frac{x_\text{gap}}{v_\text{rel}}.
\end{equation}
The time gap is the time it will take a vehicle to close the current gap with the preceding vehicle, given its current velocity, 
\begin{equation}
\label{eq:time_gap}
t_\text{gap} = \frac{x_\text{gap}}{v_\text{agent}}.
\end{equation}
In these equations, TTC is time-to-collision, $v_\text{rel}$ is the relative velocity of the agent and the preceding vehicles, and $x_\text{gap}$ is the distance gap between the vehicles. This distance gap is visualized in Figure~\ref{fig:high_d_example}. Finally, $v_\text{agent}$ is the longitudinal velocity of the agent vehicle.

Both TTC and time gap are available in the HighD dataset for human behavior; for the agent behavior, the metrics are calculated using the equations~\eqref{eq:ttc} \&~\eqref{eq:time_gap}.

Again, quantifying an acceptable error margin can only be done for a specific controller. Because we don't demonstrate a controller, we can only show the difference between the model and human behavior, but in this case study, we cannot quantify if this is acceptable for any specific IAC.

\subsection{Model training}
The optimization procedure to find the weights that fit a human demonstration best is the same as used by Sadigh et al.~\cite{Sadigh2018}. The negated log-likelihood function as proposed by Levine and Koltun~\cite{Levine2012} is minimized with respect to the weights. To keep this tractable, the human demonstration is divided into sections with the same number of frames as the control horizon used in the agent ($N=5$). All data frames are used, so the time step is $\frac{1}{25}~s$ and the planning horizon is $\frac{1}{5}~s$. The log-likelihood functions of the parts of the demonstration are summed and the summed negated log-likelihood is minimized. We assume that every lane-change trajectory in the dataset comes from a different human, an agent is trained separately for every trajectory, this resulted in $3279$ trained agents. Demonstrations on which the optimization procedure fails (i.e., no minimum of the negated log-likelihood function could be found) were discarded ($2302$ demonstrations, $41\%$). 

Because highway data is used, the velocities of the vehicles are high (mean = $29.7 m/s$) and heading angles are small. The heading angles of the vehicles are ignored in the dataset. For this reason, the dynamics of the vehicles are modeled as point masses. Because the trajectories are extracted from videos, no direct acceleration data was recorded. Acceleration data is available from the HighD dataset, but this has been reconstructed from velocity data. For this reason, the humans in the demonstrations are assumed to have direct control over the longitudinal and lateral velocities. Making the state and action vectors both 2-dimensional containing respectively an $x, y$-position and -velocity. This assumption is justified because the goal of the model is to learn the reward function, not the dynamics of human control.

\subsection{Validation of agent behavior}

To validate the agent's behavior, we evaluate the response of every agent individually in the same scenario that was used to train the agent. A dedicated test-set is not required contrary to most machine learning approaches because the log-likelihood optimization proposed by Levine and Koltun accounts for sub-optimal demonstrations by humans. This means that the learned reward function does not need to be fully optimized in the human-driven demonstration, but the agent will fully optimize the reward function. So the agent might display different behavior than the human in the same situation and thus this situation can be re-used for validation. 

For evaluation, the agent will be placed in the same initial position and its behavior is recorded for the same duration as the demonstration trajectory. Because the agent learned its reward function from this exact situation, this is a best-case scenario for the model. This approach also has the advantage that we can directly compare the agent's behavior to the human demonstration it was trained on.

As for the IRL training, heading angles are neglected and the dynamics of the vehicle are assumed to be point mass dynamics. To approximate the states and actions of real drivers, the agents are assumed to have direct control over the linear accelerations of the vehicle. This results in a 4-dimensional state vector per vehicle, containing both $x, y$-position and -velocity, and a 2-dimensional action vector containing the $x, y$- accelerations. The agent is a utility-maximizing rational agent, so it will select an action $a$ in state $s$, that maximizes its summed reward function $R$ over a time horizon $N=5$. Again, the times-step is equal to the frame rate of the HighD dataset ($\frac{1}{25}~s$). The agent has full knowledge about the future trajectories of all adjacent vehicles. As for the choice of situation, this can be regarded as a best-case scenario for the agent, since it has a perfect prediction system to predict other human behavior. 

The direct control over lateral accelerations, combined with the point mass dynamics, can result in trajectories that are not subject to normal vehicle dynamic constraints. To approximate normal vehicle dynamics, the agent's actions ($x, y$- accelerations) are constrained to the maximal values of these accelerations found in the HighD dataset. The $x$-acceleration is constrained between $(-6.63, 20.06)~m/s^2$ and the $y$-acceleration between $(-1.63, 1.63)~m/s^2$.

\section{Case study: Results}
From the first 57 recordings in the HighD dataset, all 5581 single lane changes to the left lane were automatically detected. These lane changes served as human demonstrations for the IRL-based driver model. Out of these 5581 demonstrations, 3279 resulted in a set of weights after the inverse reinforcement learning procedure. For the other $2302$ demonstrations, the IRL procedure failed to converge. 

In practice, the failure of the IRL procedure means that the likelihood function adopted from~\cite{Levine2012} becomes intractable. This function contains the logarithm of the determinant of the Hessian matrix ($log|-\textbf{H}|$). When this determinant becomes negative, the optimization fails. We found that this can happen when the optimization algorithm assigns a positive value to $\theta^h_{vel}$ (i.e., when deviating from the desired velocity is rewarded instead of punished). Note that weights are not restricted to be either positive or negative by the IRL procedure. IRL learns if a feature represents a reward or penalty for the human demonstration.

To examine if this was the cause of the high rate of failures in our training procedure, we estimated the Jacobian used in the optimization procedure for the initial values of $\theta$. For $97\%$ of the demonstrations where IRL failed, the Jacobian value for $\theta^h_{vel}$ was negative and had a magnitude at least 10 times larger than all other Jacobian values. This did not happen in demonstrations where IRL succeeded (0 out of 100 randomly selected cases). Which indicates that the optimization algorithm attempted to use positive weights for $\theta^h_{vel}$ as they have a high likelihood to explain the demonstration in the failed cases.

This could mean that features in the reward function that are based on the deviation from a maximum observed (or allowed) velocity are not suitable for use on real-world traffic conditions. On the other hand, the IRL procedure might not have failed in these cases if $\theta^h_{vel}$ was restricted to always be negative (or more generally, if weights are restricted to represent either rewards \textit{or} penalties). Further investigation to answer these questions is left for future work. We discarded the demonstrations were training failed and continued the attempt to validate the IRL-based driver model using the training data for which the model converged.

The 3279 agents that trained successfully were placed in the same scenario they were trained on to examine to what extent they show human-like behavior on a tactical and operational level. We would like to remind the reader that, combined with the fact that all agents had access to perfect predictions of all surrounding vehicles, this constituted a 'best-case scenario' for the model.

\begin{table}[h!]
    \centering
    \begin{tabular}{|c|c|c|c|}
    \hline
         & number of agents & percentage of agents & percentage of human demonstrations \\ \hline
        Lane-change & 1318 & 40.2\% & 100.0\%\\
        Collision & 875 & 26.7\% & 0.0\% \\
        Car following & 593 & 18.1\% & 0.0\%\\ 
        Off-road & 493 & 15.0\% & 0.0\%\\ \hline
        \textbf{Total} & \textbf{3279} & \textbf{100\%} & \textbf{100\%} \\ \hline
    \end{tabular}
    \caption{Tactical behavior as shown by the IRL agents and in the human-driven demonstrations the agents were trained on.}
    \label{tab:tactical_behavior}
\end{table}
\subsubsection*{Tactical behavior}
On a tactical level, we have defined four possible behaviors to categorize the resulting agent behavior: car following, lane changing, colliding, and crossing the road boundaries. Only in $40.2\%$ of the cases the model showed the same tactical behavior as the human demonstration, a lane change (Table~\ref{tab:tactical_behavior}). In more than $41\%$ of the cases, the model either collided or went on an off-road adventure. This behavior was not present in the chosen subset of the human data, so we conclude that model behavior is inconsistent with human behavior.

\begin{figure}[!h]
    \centering
    \includegraphics[width=\textwidth]{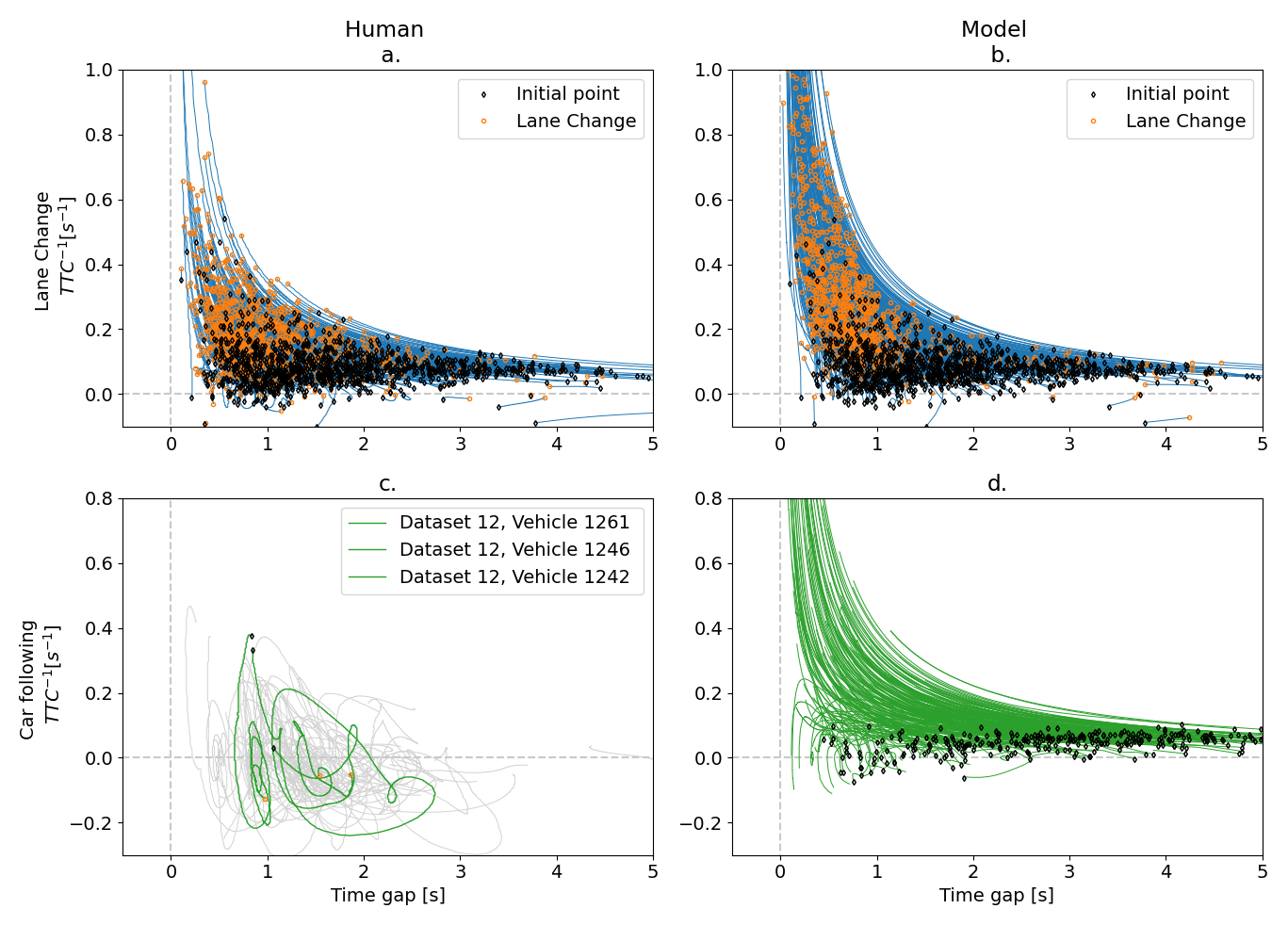}
    \caption{Inverse TTC vs time gap plots of human demonstrations (panels a and c) and IRL agent behavior (panels b and d) in lane changes and car following. Panel a shows human behavior in the used demonstrations. Since these demonstrations do not contain any car following, panel c shows 55 illustrative examples of car-following behavior selected from other trajectories in the dataset, three are highlighted for clarity. }Black diamonds indicate the initial position, this is the first frame in which a vehicle appears in the HighD dataset. In panels a and b, orange dots indicate a lane change, corresponding to the frame in which the center of a vehicle crosses the center-line between lanes. From panels a and b we conclude that the model's lane change behavior has human-like dynamics in general, however, the model makes lane changes at substantially higher inverse-TTC (lower TTC) compared to humans. From panels c and d we conclude that the model's car-following behavior does not resemble human car-following behavior.
    \label{fig:operational_results}
\end{figure}

\subsubsection*{Operational behavior}
We then compared the operational behavior of the model to the operational behavior in the human demonstrations using the inverse time to collision vs time gap plots (Figure~\ref{fig:operational_results}). Trajectories with multiple preceding vehicles show jumps in these plots due to suddenly changing values, for that reason those trajectories were omitted. Agents and humans that perform a lane change when the preceding vehicle is out of sight are also omitted since no inverse TTC and time gap data can be calculated for them for the final frames. All car-following trajectories are cropped to the point where the preceding vehicle gets out of sight. 

The plots on the left side of Figure~\ref{fig:operational_results} (a \& c) show human operational driving behavior. In the case of lane changing (\ref{fig:operational_results}-a.), the inverse TTC increases while the time gap decreases, until the point where the center lane-marking is crossed, depicted with an orange circle. In the case of car-following (\ref{fig:operational_results}-c), humans oscillate around a preferred equilibrium point. 

\begin{figure}[!h]
    \centering
    \includegraphics[width=\textwidth]{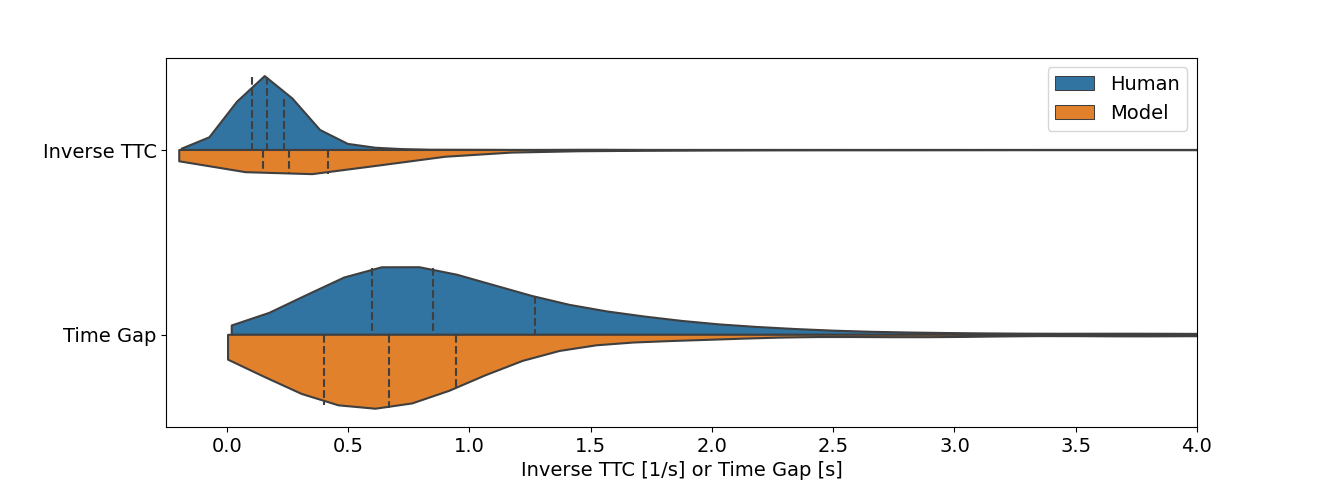}
    \caption{Estimated distributions of inverse time to collision and time gap at the moment of the lane change. The orange distributions represent the model's behavior and the blue distributions represent the human demonstrations. The mean values for inverse TTC are $0.19~s^{-1}$ for human lane changes and $0.47~s^{-1}$ for the model. The mean values for time gap are $1.05~s$ for human behavior and $0.85~s$ for the model behavior.}
    \label{fig:violin}
\end{figure}

The model's behavior for the same maneuvers can be seen on the right side of Figure~\ref{fig:operational_results}. The model's lane-changing behavior (\ref{fig:operational_results}-b) has human-like dynamics in general (as in \ref{fig:operational_results}-a), however, the model makes lane changes at substantially higher inverse TTC (lower TTC) compared to humans. Also, the time gap at the moment of the lane change is on average smaller than for the human demonstrations. To further illustrate the differences in the lane change dynamics, we investigated the distributions of inverse TTC and time gap at the moment of lane change (Figure~\ref{fig:violin}). This shows substantial differences between the estimated distributions. We performed a paired t-test to check for significant differences, both the inverse TTC ($t(1075)=-7.61$, $p=6.1\mathrm{e}{-14}<0.001$, Cohen d=$0.302$) and time gap ($t(1075)=13.49$, $p=2.0\mathrm{e}{-38}<0.001$, Cohen d=$0.234$) values at the moment of lane change differ significantly between the model and human demonstrations. So for lane-changing behavior, we conclude that the IRL-based model does not resemble human behavior on an operational level.

When comparing the agent's car-following behavior (\ref{fig:operational_results}-d) with the human's car-following behavior (\ref{fig:operational_results}-c), there are no oscillations around an equilibrium point for most agents. The general shape resembles that of a human lane-changing maneuver (\ref{fig:operational_results}-a) without crossing the center lane-marking. From this, we conclude that if the model shows car-following behavior, it does not do that in a way that resembles human oscillatory car-following behavior but instead it tailgates the preceding vehicle. 

\subsubsection*{Reason for the agents' behavior}
Why do the IRL-based agents show behavior that is so different from human behavior, even though their reward function was learned from human demonstrations? We randomly selected several agent trajectories for manual examination using the TraViA~\cite{Siebinga2021} traffic visualization tool to answer this question. Examples of these trajectories can also be found as videos in the supplementary materials. From these manual evaluations, two main causes were identified that explain why the behavior of the agents does not represent human behavior: the model's assumptions and the IRL fitting procedure.

To start with the cases where the model's assumptions cannot explain the desired behavior. Consider a demonstration where a human is merging in a slow-moving and crowded left lane to overtake a truck farther ahead in the right lane. This might be beneficial in the long run because the truck can be overtaken, but such behavior is unlikely to be beneficial within the short planning horizon of the model, especially because the distance-based collision features promote staying away from other vehicles. This issue is similar to the previously identified problem that lane changes to a right lane cannot be explained by the currently assumed reward function. In both of these cases, no matter the learned weights, the assumed reward function will not lead to the desired behavior within the planning horizon.

\begin{figure}[h!]
    \centering
    \includegraphics[width=\textwidth]{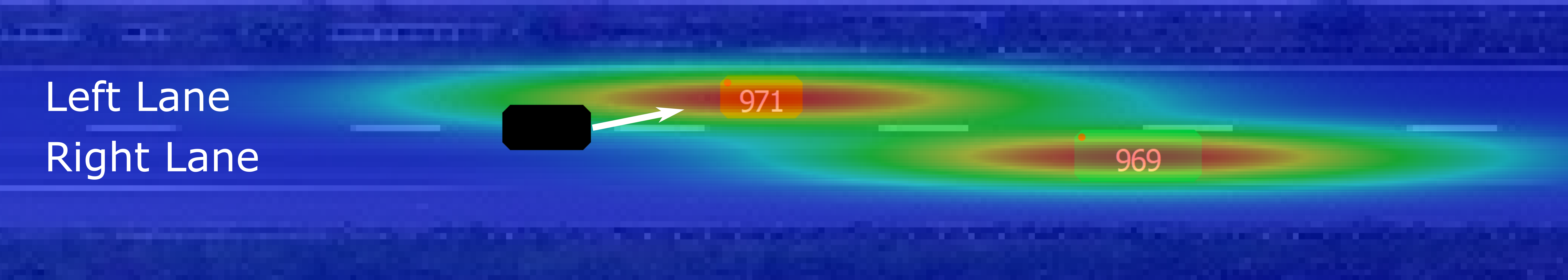}
    \caption{An example of a demonstration where the assumed anti-collision feature does not describe the human's demonstrated behavior. In this figure, the black shape represents the position of the human-driven demonstration vehicle during a lane change maneuver, the white arrow indicates its direction of motion. Only the collision feature is visualized with warmer colors indicating higher cost. In this example, the demonstrating vehicle is moving from a low-cost area (right lane) to a high-cost area (left lane). The only way to explain this behavior with the assumed features is to assign a low weight to the collision feature. Apparently, the demonstrating human does not care so much about moving into the higher cost area in the left lane, other features must be more important. When these learned weights are then used in a utility-maximizing agent in the same situation, it will not make the demonstrated lane change. Instead, it will stay in the right lane with a lower penalty and finally collide with the preceding vehicle (969), because collision prevention has a low weight.}
    \label{fig:wrong_learning}
\end{figure}

In other cases, the approach of learning the weights from a demonstration using an assumed reward function can be identified as the cause of the problem. Many agents that collided learned their weights from a demonstration where the human moves into the area influenced by the collision feature (see Figure~\ref{fig:wrong_learning} for an example). Because the dimensions of this collision feature are fixed and only the weights are learned in the IRL procedure, the resulting collision weight will be low, i.e. a low collision weight is the only way to explain the human moving into this area. When this low collision weight is used in the agent to generate behavior, the agent will not perform a lane change, because moving into the collision-feature area will always decrease the reward. Instead, the agent will stay in its lane. When it approaches the preceding vehicle, it will collide, because of the low collision-prevention weight. 

The underlying problem here is that the assumed reward function cannot describe the human's demonstrated behavior properly. Suppose using such a flawed reward function with hand-picked weights. In that case, one would expect prediction errors on the operational level, because the timing of the lane change is determined by the distance-based collision feature. In this case however, the IRL procedure exaggerates the effects of the flawed reward function by learning weights that result in more collisions. So even though the problem lies in the flawed reward function and not the IRL procedure itself, the combination of the IRL-procedure and reward function might not only limit the performance of the model, it can actively make it worse.

\section{Discussion}
In this work, we have proposed a validation workflow for driver models in interaction-aware AV controllers. We illustrated its utility through a case study of validating an inverse reinforcement learning-based driver model replicated from literature~\cite{Sadigh2018} using naturalistic highway driving data extracted from the HighD dataset~\cite{Krajewski2018}. Our validation workflow (Figure~\ref{fig:proposed_workflow}) incorporates the automatic extraction of comparable lane change scenarios (5581) on which the IRL model was trained (step 1). The validation of the model was then performed in two related stages. First, we examined the tactical behavior of the model (step 2). Even though no collisions or off-road driving were present in the training data, the model produced such behavior in more than $41\%$ of the cases (Table~\ref{tab:tactical_behavior}). Second, we analyzed the operational behavior of the model in the $59\%$ remaining trajectories (step 3). This analysis revealed that even though the dynamics of the model's lane changes are similar to humans (Figure ~\ref{fig:operational_results}a,b), the model performed the lane changes with significantly smaller safety margins (Figure~\ref{fig:violin}). Furthermore, the dynamics of the model's car following behavior was largely inconsistent with human behavior (Figure~\ref{fig:operational_results}c,d). 

In conclusion, despite training the IRL-based model on data of real-world driving behavior, our 3-step evaluation workflow exposed how the model is not able to produce realistic behavior in the same scenarios. This case study illustrated that despite promising results in simple IAC demonstrations, the models used for human behavior prediction in IACs can deviate substantially from actual human behavior, which can have serious ramifications for generalization of IACs to real-world environments. Our results highlight the importance of validating the models used in interaction-aware motion planning for autonomous vehicles, and suggest an easy-to-use framework to aid researchers in doing so.

\subsubsection*{Practical applicability of the validation workflow}

The case study of validating an IRL-based driver model illustrated the practical applicability of the proposed validation workflow (Figure~\ref{fig:proposed_workflow}). In the first step of the workflow, the case study showed the feasibility of automatically extracting data from an open-access naturalistic dataset. Even after narrowing down the extracted data to select specific scenarios (in our case, lane changes), the data were sufficiently rich to serve as training data for the IRL model. Note that multiple other datasets were available for consideration (e.g., NGSim~\cite{NGSIM2016} \& PNeuma~\cite{Barmpounakis2020}) to further enlarge the data and/or use scenarios other than lane changes. 

The second and third steps of the workflow propose a two-stage evaluation approach, separated into \textit{tactical} and \textit{operational} driver behavior. The case study illustrated why this two-stage validation is useful and necessary. On the tactical level, the large number of collisions and off-road driving would have been hard to identify in a one-stage metric-based validation (e.g. mean square-root error in~\cite{Schwarting2019a}). On the operational level, the evaluation illustrated that the differences between car following and lane changing in human behavior were not reflected in the model's behavior. This would have been impossible to identify without first examining the tactical behavior.

The results of the case study also underline the importance of validating driver models for IACs in general. The discrepancy between the driver model and human behavior suggests that an IAC using this model might not safely generalize to real-world scenarios. The case study shows that models that do not actually capture human behavior are not just a hypothetical issue, but a practical concern for IACs developed for autonomous vehicles.

\subsubsection*{Implications for interaction-aware controllers}
The results of the IRL-based model validation have implications for IACs that would use this model to predict other drivers' responses. Wrong predictions on the tactical level can lead to dangerous situations. If an AV decides to accelerate based on an inaccurate prediction that a vehicle in an adjacent lane will stay there, a dangerous situation might occur when the other vehicle moves in front of the AV. The same holds for inaccurate predictions on an operational level. For example, the model will close the gaps to a (very) high inverse TTC (low TTC) compared to human drivers. This can lead to over-conservative AV behavior because the controller over-estimates the aggressiveness of the human. The full extent of these implications needs to be further examined in future work. 

\subsubsection*{Related work and generalizability}
To the best of our knowledge, our work is the first attempt to validate a driver model used in interaction-aware controllers on both the tactical and operational levels. The work on which this model was based~\cite{Sadigh2018} does not use naturalistic data and reports no validation attempt of the behavior model. Another related study~\cite{Schwarting2019} does use naturalistic data (NGSim) to train the IRL-based model, but also does not report any validation of the trained model. In the supplementary material (available at~\cite{Schwarting2019a}) Schwarting et al. do report the mean squared errors of their model for merging scenarios. However, given the complexity of human behavior in traffic interactions, such one-dimensional averaged error metrics provide only rudimentary information on how well the model captures human behavior. 

Driver model validation using naturalistic data has been performed for other use cases than IACs. In~\cite{Zhu2018} five car-following models are validated for use in microscopic traffic simulations on naturalistic data collected in Shanghai. Their validation method could also be useful when designing IACs, and our validation method could as well be used to validate models developed for applications other than IACs. However, we argue that because our method includes the tactical and operational validation steps, it is more suitable to validate models displaying multiple higher-level behaviors.

Besides the IAC literature, there have been other driver modeling attempts using IRL. However, IRL-based models can differ substantially from each other in terms of the used reward function. Naumann et al. studied the suitability of different cost functions for different driving scenarios~\cite{Naumann2020} and showed that there are substantial differences. The other modeling attempts that use IRL differ from our work precisely in the sense that they target another scenario (e.g.~\cite{Rosbach2019} who regards curve negotiation) or use different reward function features (e.g.~\cite{Huang2021} who uses velocity-based features for risk perception). That means that the results of those works should be regarded as validations of different models, despite the fact that they are also based on IRL. This observation leads to two conclusions. First, other models that use different features should be validated as new models even when they also use IRL. Second, the choice of features for the reward function impacts the performance of the model, which might provide an opportunity to improve models that underperform. 

The reasons why the IRL-based models perform poorly in our case study will most likely generalize to other IRL-based models that use similar distance-based collision features in the reward function. The results show that such distance-based features do not capture the essence of human driving behavior. Only changing the shape or dimension of a position-based feature will not solve this. Instead, we advocate that the metrics used in the reward function features should be based on human factors literature for the targeted tactical behaviors, as was done in the operational validation of the model; e.g. the distance-based collision feature could be replaced with a TTC-based feature (similar to the previously mentioned model in~\cite{Huang2021}). 

Other validation attempts of human driver models that do not specifically target IACs and do not use IRL also exist, one especially related to our work is~\cite{Srinivasan2021}. In that work, Srinivasan et al. compare the trajectories generated with a deep-learning-based model to naturalistic driving data. As in our work, the comparison is based on an in-depth analysis of the resulting trajectories instead of one-dimensional metrics. They show that, also for deep-learning-based driver models, validation should be grounded on a low-level comparison of trajectories, not just high-level metrics. They do however not provide a generalized framework for performing such validations as we do with our proposed workflow.

\subsubsection*{Limitations and recommendations}
This work has three main limitations. First, we used only a single demonstration of a lane change to train the IRL, which might explain part of the discrepancy between human and model behavior in the results. However, providing the system with more training data might only slightly improve the model's performance. In the case study we identified the causes of the observed problems to be the features of the reward function, not the weights. Adding more training data could result in weights that better fit a specific driver. But it will not negate the problem with the features used in the reward function. 

Second, it should be noted that the planning horizon of the model is very short due to the combination of a low number of frames and a high frame rate ($N=5$ at $25~Hz$). The number of frames within the planning horizon was chosen based on the previous work~\cite{Sadigh2018} and to keep the IRL procedure tractable. The frame rate was directly adopted from the HighD dataset for simplicity, both for reproduction purposes and to not introduce any extra assumptions when down-sampling the data. Increasing the planning horizon and examining the model's behavior under those conditions is left for future work.

Finally, our case study only attempts to validate the model, it does not quantify the implications of the outcome for use of the model in an IAC. Therefore, we are unable to say which aspects of the model's behavior would be tolerable when used in an IAC or which aspects have major consequences. Quantifying the implications of the mismatch between the model's, and naturalistic human behavior is left for future work. Answering such a question is an interesting topic of research on its own, a perspective on how to approach such an evaluation can be found in~\cite{Markkula2022}.

Future work should also focus on validating more driver models for use in AV controllers, e.g. the Intelligent Driver Model~\cite{Treiber2000} mentioned in the introduction is used in many simulations and demonstrations to model individual human behavior for IACs and should be validated for such use. Future work on IRL-based driver models could focus on redesigning the used reward function such that it better captures similarities between human drivers by using human-factors literature as a starting point. Besides that, the IRL-based model used here could be extended to take the uncertainty in human behavior into account. Either the uncertainty over the learned rewards could be targeted by learning multiple reward functions (as is done in~\cite{Naumann2020, Sun2020}) instead of only single parameters and selecting the best fit, or stochasticity could be added when selecting the actions to relax the assumption of humans being utility maximizers (also done in~\cite{Sun2020}). However, such changes to the model could complicate the implementation in an IAC.

\section{Conclusions}
In this paper, we argued for validation of the driver models used in interaction-aware controllers. We proposed an evaluation workflow for such validation, illustrated through a concrete case study. Based on the findings in our paper, we conclude the following:
\begin{itemize}
    \item The proposed workflow allowed for a detailed evaluation of a driver model replicated from literature, based on an open-source dataset from which 3279 human-driven lane changes in moderately heavy highway traffic could be extracted. After training the model on each lane change, it did not reproduce adequate behavior when exposed to the same conditions. It generated crashes and road departures in 41.7\% of the cases (inadequate tactical behavior). For the remaining cases, unrealistic safety margins were observed (inadequate operational behavior). These unrealistic predictions show that models that do not capture realistic human behavior are a practical concern for implementing IACs in future autonomous vehicles.
    \item During the case study, the proposed workflow proved to be practically applicable, providing a structured basis for model validation in two stages:
    \begin{itemize}
        \item First, validating the tactical behavior illustrated to what extent high-level choices are correctly predicted (e.g., that a lane change occurs, rather than staying behind the lead vehicle see Table~\ref{tab:tactical_behavior}).
        \item Second, correct tactical behaviors produced by a model should be validated in additional detail, by evaluating to what extent the behavior is executed in a way that resembles the timing and spatio-temporal safety margins acceptable to human drivers (see Figure~\ref{fig:operational_results}).
        \item In these two stages, different tactical behaviors should be evaluated based on different operational criteria because differences in human operational behavior were observed for different tactical behaviors (see Figure~\ref{fig:operational_results}).
    \end{itemize}
\end{itemize}

\section*{Acknowledgements}
The authors thank Nissan Motor Co. Ltd. for funding this work and we especially thank Daniel Góngora for his valuable feedback and contributions to the article.

\bibliographystyle{ieeetr}
\bibliography{My_Collection}

\begin{thebibliography}{10}

\bibitem{Harper2016}
C.~D. Harper, C.~T. Hendrickson, S.~Mangones, and C.~Samaras, ``{Estimating
  potential increases in travel with autonomous vehicles for the non-driving,
  elderly and people with travel-restrictive medical conditions},'' {\em
  Transportation Research Part C: Emerging Technologies}, vol.~72, pp.~1--9,
  nov 2016.

\bibitem{Clements2017}
L.~M. Clements and K.~M. Kockelman, ``{Economic Effects of Automated
  Vehicles},'' {\em Transportation Research Record: Journal of the
  Transportation Research Board}, vol.~2606, pp.~106--114, jan 2017.

\bibitem{Pettigrew2017}
S.~Pettigrew, ``{Why public health should embrace the autonomous car},'' {\em
  Australian and New Zealand Journal of Public Health}, vol.~41, pp.~5--7, feb
  2017.

\bibitem{Sadigh2018}
D.~Sadigh, N.~Landolfi, S.~S. Sastry, S.~A. Seshia, and A.~D. Dragan,
  ``{Planning for cars that coordinate with people: leveraging effects on human
  actions for planning and active information gathering over human internal
  state},'' {\em Autonomous Robots}, vol.~42, pp.~1405--1426, oct 2018.

\bibitem{Schwarting2019}
W.~Schwarting, A.~Pierson, J.~Alonso-Mora, S.~Karaman, and D.~Rus, ``{Social
  behavior for autonomous vehicles},'' {\em Proceedings of the National Academy
  of Sciences}, vol.~116, pp.~24972--24978, dec 2019.

\bibitem{Evestedt2016}
N.~Evestedt, E.~Ward, J.~Folkesson, and D.~Axehill, ``{Interaction aware
  trajectory planning for merge scenarios in congested traffic situations},''
  {\em IEEE Conference on Intelligent Transportation Systems, Proceedings,
  ITSC}, pp.~465--472, 2016.

\bibitem{Ward2017}
E.~Ward, N.~Evestedt, D.~Axehill, and J.~Folkesson, ``{Probabilistic Model for
  Interaction Aware Planning in Merge Scenarios},'' {\em IEEE Transactions on
  Intelligent Vehicles}, vol.~2, no.~2, pp.~1--1, 2017.

\bibitem{Liu2015}
W.~Liu, S.-W. Kim, S.~Pendleton, and M.~H. Ang, ``{Situation-aware decision
  making for autonomous driving on urban road using online POMDP},'' in {\em
  2015 IEEE Intelligent Vehicles Symposium (IV)}, vol.~2015-Augus,
  pp.~1126--1133, IEEE, jun 2015.

\bibitem{Meng2016}
F.~Meng, J.~Su, C.~Liu, and W.-H. Chen, ``{Dynamic decision making in lane
  change: Game theory with receding horizon},'' in {\em 2016 UKACC 11th
  International Conference on Control (CONTROL)}, pp.~1--6, IEEE, aug 2016.

\bibitem{Lenz2016}
D.~Lenz, T.~Kessler, and A.~Knoll, ``{Tactical cooperative planning for
  autonomous highway driving using Monte-Carlo Tree Search},'' in {\em 2016
  IEEE Intelligent Vehicles Symposium (IV)}, vol.~2016-Augus, pp.~447--453,
  IEEE, jun 2016.

\bibitem{Tian2019}
R.~Tian, S.~Li, N.~Li, I.~Kolmanovsky, A.~Girard, and Y.~Yildiz, ``{Adaptive
  Game-Theoretic Decision Making for Autonomous Vehicle Control at
  Roundabouts},'' in {\em 2018 IEEE Conference on Decision and Control (CDC)},
  vol.~2018-Decem, pp.~321--326, IEEE, dec 2018.

\bibitem{Zhang2018}
Q.~Zhang, D.~Filev, H.~E. Tseng, S.~Szwabowski, and R.~Langari, ``{Addressing
  Mandatory Lane Change Problem with Game Theoretic Model Predictive Control
  and Fuzzy Markov Chain},'' in {\em 2018 Annual American Control Conference
  (ACC)}, vol.~2018-June, pp.~4764--4771, IEEE, jun 2018.

\bibitem{Yu2018}
H.~Yu, H.~E. Tseng, and R.~Langari, ``{A human-like game theory-based
  controller for automatic lane changing},'' {\em Transportation Research Part
  C: Emerging Technologies}, vol.~88, pp.~140--158, mar 2018.

\bibitem{Hubmann2018}
C.~Hubmann, J.~Schulz, G.~Xu, D.~Althoff, and C.~Stiller, ``{A Belief State
  Planner for Interactive Merge Maneuvers in Congested Traffic},'' in {\em 2018
  21st International Conference on Intelligent Transportation Systems (ITSC)},
  vol.~2018-Novem, pp.~1617--1624, IEEE, nov 2018.

\bibitem{Coskun2019}
S.~Coskun, Q.~Zhang, and R.~Langari, ``{Receding Horizon Markov Game Autonomous
  Driving Strategy},'' in {\em 2019 American Control Conference (ACC)},
  vol.~2019-July, pp.~1367--1374, IEEE, jul 2019.

\bibitem{Garzon2019}
M.~Garz{\'{o}}n and A.~Spalanzani, ``{Game theoretic decision making for
  autonomous vehicles' merge manoeuvre in high traffic scenarios},'' {\em 2019
  IEEE Intelligent Transportation Systems Conference, ITSC 2019},
  pp.~3448--3453, 2019.

\bibitem{Isele2019}
D.~Isele, ``{Interactive Decision Making for Autonomous Vehicles in Dense
  Traffic},'' in {\em 2019 IEEE Intelligent Transportation Systems Conference
  (ITSC)}, pp.~3981--3986, IEEE, oct 2019.

\bibitem{Hang2021}
P.~Hang, C.~Lv, Y.~Xing, C.~Huang, and Z.~Hu, ``Human-like decision making for
  autonomous driving: A noncooperative game theoretic approach,'' {\em IEEE
  Transactions on Intelligent Transportation Systems}, vol.~22, no.~4,
  pp.~2076--2087, 2021.

\bibitem{Michon1985}
J.~A. Michon, ``{A Critical View of Driver Behavior Models: What Do We Know,
  What Should We Do?},'' in {\em Human Behavior and Traffic Safety},
  pp.~485--524, Boston, MA: Springer US, 1985.

\bibitem{Ranney2011}
T.~A. Ranney, ``{Psychological fidelity: Perception of risk},'' in {\em
  Handbook of Driving Simulation for Engineering, Medicine, and Psychology}
  (D.~L. Fisher, M.~Rizzo, J.~Caird, and J.~D. Lee, eds.), ch.~9, pp.~9--1
  --9--13, 2011.

\bibitem{Lee1976}
D.~N. Lee, ``{A Theory of Visual Control of Braking Based on Information about
  Time-to-Collision},'' {\em Perception}, vol.~5, pp.~437--459, dec 1976.

\bibitem{Greenberg2011}
J.~Greenberg and M.~Blommer, ``{Physical Fidelity of Driving Simulators},'' in
  {\em Handbook of Driving Simulation for Engineering, Medicine, and
  Psychology}, no.~2011, pp.~7--1--7--24, CRC Press, apr 2011.

\bibitem{Treiber2000}
M.~Treiber, A.~Hennecke, and D.~Helbing, ``{Congested traffic states in
  empirical observations and microscopic simulations},'' {\em Physical Review E
  - Statistical Physics, Plasmas, Fluids, and Related Interdisciplinary
  Topics}, vol.~62, no.~2, pp.~1805--1824, 2000.

\bibitem{Antin2019}
J.~F. Antin, S.~Lee, M.~A. Perez, T.~A. Dingus, J.~M. Hankey, and A.~Brach,
  ``{Second strategic highway research program naturalistic driving study
  methods},'' {\em Safety Science}, vol.~119, pp.~2--10, nov 2019.

\bibitem{Krajewski2018}
R.~Krajewski, J.~Bock, L.~Kloeker, and L.~Eckstein, ``{The highD Dataset: A
  Drone Dataset of Naturalistic Vehicle Trajectories on German Highways for
  Validation of Highly Automated Driving Systems},'' in {\em 2018 21st
  International Conference on Intelligent Transportation Systems (ITSC)},
  vol.~2018-Novem, pp.~2118--2125, IEEE, nov 2018.

\bibitem{NGSIM2016}
{U.S. Department of Transportation Federal Highway Administration}, ``{Next
  Generation Simulation (NGSIM) Vehicle Trajectories and Supporting Data.
  [Dataset]},'' 2016.

\bibitem{Barmpounakis2020}
E.~Barmpounakis and N.~Geroliminis, ``{On the new era of urban traffic
  monitoring with massive drone data: The pNEUMA large-scale field
  experiment},'' {\em Transportation Research Part C: Emerging Technologies},
  vol.~111, no.~November 2019, pp.~50--71, 2020.

\bibitem{Siebinga2021}
O.~Siebinga, ``{TraViA: a Traffic data Visualization and Annotation tool in
  Python},'' {\em Journal of Open Source Software}, vol.~6, p.~3607, sep 2021.

\bibitem{Saifuzzaman2014}
M.~Saifuzzaman and Z.~Zheng, ``{Incorporating human-factors in car-following
  models: A review of recent developments and research needs},'' {\em
  Transportation Research Part C: Emerging Technologies}, vol.~48,
  pp.~379--403, nov 2014.

\bibitem{Ossen2011}
S.~Ossen and S.~P. Hoogendoorn, ``{Heterogeneity in car-following behavior:
  Theory and empirics},'' {\em Transportation Research Part C: Emerging
  Technologies}, vol.~19, no.~2, pp.~182--195, 2011.

\bibitem{Hoogendoorn2006}
S.~Hoogendoorn, S.~Ossen, and M.~Schreuder, ``{Empirics of Multianticipative
  Car-Following Behavior},'' {\em Transportation Research Record: Journal of
  the Transportation Research Board}, vol.~1965, pp.~112--120, jan 2006.

\bibitem{Jiang2015}
R.~Jiang, M.~B. Hu, H.~M. Zhang, Z.~Y. Gao, B.~Jia, and Q.~S. Wu, ``{On some
  experimental features of car-following behavior and how to model them},''
  {\em Transportation Research Part B: Methodological}, vol.~80, pp.~338--354,
  2015.

\bibitem{Mulder2005}
M.~Mulder, M.~Mulder, M.~M. {Van Paassen}, and D.~A. Abbink, ``{Effects of lead
  vehicle speed and separation distance on driver car-following behavior},''
  {\em Conference Proceedings - IEEE International Conference on Systems, Man
  and Cybernetics}, vol.~1, no.~4, pp.~399--404, 2005.

\bibitem{KONDOH2008}
T.~KONDOH, T.~YAMAMURA, S.~KITAZAKI, N.~KUGE, and E.~R. BOER, ``{Identification
  of Visual Cues and Quantification of Drivers' Perception of Proximity Risk to
  the Lead Vehicle in Car-Following Situations},'' {\em Journal of Mechanical
  Systems for Transportation and Logistics}, vol.~1, no.~2, pp.~170--180, 2008.

\bibitem{Ng2000}
A.~Ng and S.~Russell, ``{Algorithms for inverse reinforcement learning},'' {\em
  Proceedings of the Seventeenth International Conference on Machine Learning},
  vol.~0, pp.~663--670, 2000.

\bibitem{Abbeel2004}
P.~Abbeel and A.~Y. Ng, ``{Apprenticeship learning via inverse reinforcement
  learning},'' {\em Proceedings, Twenty-First International Conference on
  Machine Learning, ICML 2004}, pp.~1--8, 2004.

\bibitem{Ziebart2008}
B.~D. Ziebart, A.~Maas, J.~A. Bagnell, and A.~K. Dey, ``{Maximum entropy
  inverse reinforcement learning},'' in {\em Proceedings of the National
  Conference on Artificial Intelligence}, vol.~3, pp.~1433--1438, 2008.

\bibitem{Siebinga2021b}
O.~Siebinga, ``{IRL Model Validation - TraViA extension code}.''
  https://github.com/tud-hri/irlmodelvalidation, 2021.

\bibitem{Mulder2009}
M.~Mulder, M.~M. {Van Paassen}, M.~Mulder, J.~J. Pauwelussen, and D.~A. Abbink,
  ``{Haptic car-following support with deceleration control},'' {\em Conference
  Proceedings - IEEE International Conference on Systems, Man and Cybernetics},
  no.~October, pp.~1686--1691, 2009.

\bibitem{Dang2013}
{Ruina Dang}, {Fang Zhang}, {Jianqiang Wang}, {Shichun Yi}, and {Keqiang Li},
  ``{Analysis of Chinese driver's lane change characteristic based on real
  vehicle tests in highway},'' in {\em 16th International IEEE Conference on
  Intelligent Transportation Systems (ITSC 2013)}, vol.~100084, pp.~1917--1922,
  IEEE, oct 2013.

\bibitem{Levine2012}
S.~Levine and V.~Koltun, ``{Continuous Inverse Optimal Control with Locally
  Optimal Examples},'' {\em Proceedings of the 29th International Conference on
  Machine Learning, ICML 2012}, vol.~1, pp.~41--48, jun 2012.

\bibitem{Schwarting2019a}
W.~Schwarting, A.~Pierson, J.~Alonso-Mora, S.~Karaman, and D.~Rus, ``{Social
  behavior for autonomous vehicles - Supporting Information},'' 2019.

\bibitem{Zhu2018}
M.~Zhu, X.~Wang, A.~Tarko, and S.~Fang, ``{Modeling car-following behavior on
  urban expressways in Shanghai: A naturalistic driving study},'' {\em
  Transportation Research Part C: Emerging Technologies}, vol.~93,
  pp.~425--445, aug 2018.

\bibitem{Naumann2020}
M.~Naumann, L.~Sun, W.~Zhan, and M.~Tomizuka, ``{Analyzing the Suitability of
  Cost Functions for Explaining and Imitating Human Driving Behavior based on
  Inverse Reinforcement Learning},'' in {\em 2020 IEEE International Conference
  on Robotics and Automation (ICRA)}, pp.~5481--5487, IEEE, may 2020.

\bibitem{Rosbach2019}
S.~Rosbach, V.~James, S.~Grosjohann, S.~Homoceanu, and S.~Roth, ``{Driving with
  Style: Inverse Reinforcement Learning in General-Purpose Planning for
  Automated Driving},'' in {\em 2019 IEEE/RSJ International Conference on
  Intelligent Robots and Systems (IROS)}, no.~November, pp.~2658--2665, IEEE,
  nov 2019.

\bibitem{Huang2021}
Z.~Huang, J.~Wu, and C.~Lv, ``Driving behavior modeling using naturalistic
  human driving data with inverse reinforcement learning,'' {\em IEEE
  Transactions on Intelligent Transportation Systems}, pp.~1--13, 2021.

\bibitem{Srinivasan2021}
A.~R. Srinivasan, M.~Hasan, Y.-S. Lin, M.~Leonetti, J.~Billington, R.~Romano,
  and G.~Markkula, ``Comparing merging behaviors observed in naturalistic data
  with behaviors generated by a machine learned model,'' 2021.

\bibitem{Markkula2022}
G.~Markkula and M.~Dogar, ``{How accurate models of human behavior are needed
  for human-robot interaction? For automated driving?},'' feb 2022.

\bibitem{Sun2020}
L.~Sun, Z.~Wu, H.~Ma, and M.~Tomizuka, ``{Expressing Diverse Human Driving
  Behavior with Probabilistic Rewards and Online Inference},'' in {\em 2020
  IEEE/RSJ International Conference on Intelligent Robots and Systems (IROS)},
  pp.~2020--2026, IEEE, oct 2020.

\end{thebibliography}

\end{document}